\documentclass[11pt,a4paper]{article}
\usepackage[hyperref]{emnlp2020}
\usepackage{times}
\usepackage{latexsym}
\usepackage{xcolor,soul}
\usepackage{multicol}
\sethlcolor{pink}

\usepackage{microtype}
\usepackage{enumerate}
\usepackage{amssymb}
\usepackage{amsmath}
\usepackage{latexsym}
\usepackage{url}
\usepackage{mathtools}
\usepackage{multirow}
\usepackage{xspace}
\usepackage{comment}
\usepackage{flowchart}
\usetikzlibrary{arrows}
\usepackage{booktabs}
\usepackage{subcaption}
\usepackage{enumitem}
\usepackage{multirow}
\usepackage{tikz}
\usepackage{pgfplots}

\usepackage{float}
\usepackage{multicol}
\restylefloat{table}
\usepackage{amsmath}
\usepackage{graphicx}
\usepackage{color}
 
\usepackage{comment}
\usepackage{enumerate}
\usepackage{enumitem}

\usepackage[nameinlink]{cleveref}
\crefformat{section}{\S#2#1#3} 
\crefformat{subsection}{\S#2#1#3}
\crefformat{subsubsection}{\S#2#1#3}

\usepackage{xcolor}
\usepackage[linesnumbered, algo2e, vlined, ruled]{algorithm2e}

\SetCommentSty{mycommfont}

\DeclareMathAlphabet{\pazocal}{OMS}{zplm}{m}{n}
\DeclareMathAlphabet{\pazocal}{OMS}{zplm}{m}{n}

\usepackage{cleveref}

\crefformat{section}{\S#2#1#3} 
\crefformat{subsection}{\S#2#1#3}
\crefformat{subsubsection}{\S#2#1#3}

\aclfinalcopy 
\def\aclpaperid{865} 

\newcommand{\red}[1]{\textcolor{black}{#1}}
\newcommand{\blue}[1]{\textcolor{black}{#1}}

\newcommand{\ie}{{\em i.e.,}\xspace}
\newcommand{\eg}{{\em e.g.,}\xspace}

\newcommand{\Ms}{\pazocal{M}}
\newcommand{\Ns}{\pazocal{N}}
\newcommand{\Vs}{\pazocal{V}}

\newcommand{\Ds}{\pazocal{D}}
\newcommand{\Es}{\pazocal{E}}

\newcommand{\Ni}{{\em i.}~}
\newcommand{\Nii}{{\em ii.}~}
\newcommand{\Niii}{{\em iii.}~}

\newcommand{\Na}{({\em a})~}
\newcommand{\Nb}{({\em b})~}
\newcommand{\Nc}{({\em c})~}
\newcommand{\Nd}{({\em d})~}
\newcommand{\Ne}{({\em e})~}
\newcommand{\Nf}{({\em f})~}

\newcommand{\lnmap}{{\sc{LNMap}}}
\newcommand{\lnmapb}{{\sc{\textbf{LNMap}}}}
\newcommand{\lnmaplin}{{\sc{{LNMap}}\ {(Lin. AE)}}}
\newcommand{\lnmaplinb}{{\sc{\textbf{LNMap}}\ \textbf{(Lin. AE)}}}

\title{{\sc{LNMap}}: Departures from Isomorphic Assumption in Bilingual Lexicon Induction Through Non-Linear Mapping in Latent Space}

\author{Tasnim Mohiuddin$^\P$, M Saiful Bari$^\P$, \and Shafiq Joty$^\P$$^\dagger$ \\
$^\P$Nanyang Technological University, Singapore \\
$^\dagger$Salesforce Research\\
{\tt \{mohi0004, bari0001, srjoty\}@ntu.edu.sg}
\\}

\date{}

\begin{document}
\maketitle
\begin{abstract}
\blue{Most of the successful and predominant methods for Bilingual Lexicon Induction (BLI) are mapping-based, where a linear mapping function is learned with the assumption that the word embedding spaces of different languages exhibit similar geometric structures (\ie\ approximately \emph{isomorphic}).} However, several recent studies have criticized this simplified assumption showing that it does not hold in general even for closely related languages.  In this work, we propose a novel semi-supervised method to \blue{learn cross-lingual word embeddings for BLI.} Our model is independent of the isomorphic assumption and uses non-linear mapping in the latent space of two \blue{independently pre-trained} autoencoders. Through extensive experiments on fifteen (15) different language pairs (in both directions) comprising resource-rich and low-resource languages from two different datasets, we demonstrate that our method outperforms existing models by a good margin. Ablation studies show the importance of different model components and the necessity of non-linear mapping.
\end{abstract}

\section{Introduction} 
\label{sec:intro}

In recent years, a plethora of methods have been proposed to learn cross-lingual word embeddings (or {CLWE} for short) from monolingual word embeddings. Here words with similar meanings in different languages are represented by similar vectors, regardless of their actual language. CLWE enable us to compare the meaning of words across languages, which is key to most multi-lingual applications such as bilingual lexicon induction \cite{heyman-17}, machine translation \cite{lample-18-phrase, artetxe-18-iclr}, or multi-lingual information retrieval \cite{vulic-15-ir}.
They also play a crucial role in cross-lingual knowledge transfer between languages (\eg\ from resource-rich to low-resource languages) by providing a common representation space \cite{ruder-19-tutorial}.

\newcite{Mikolov13}, in their pioneering work, learn a \textit{linear} mapping function to transform the source embedding space to the target language by minimizing the squared Euclidean distance between the translation pairs of a seed dictionary. They assume that the {similarity of geometric arrangements in the embedding spaces} is the key reason for their method to succeed as they found linear mapping superior to non-linear mappings with multi-layer neural networks. Subsequent studies propose to improve the model by normalizing the embeddings, imposing an {orthogonality constraint} on the linear mapper, modifying the objective function, and reducing the seed dictionary size \cite{artetxe2016emnlp,artetxe2017acl,artetxe2018aaai,SmithICLR17}. 




A more recent line of research attempts to eliminate the seed dictionary totally and learn the mapping in a purely unsupervised way \cite{Valerio16, Zhang17, conneau2018word, Artetxe-2018-acl, Xu2018, Hoshen-18, david2018gromov, mohi-joty:2019, mohi-joty-cl}. While not requiring any cross-lingual supervision makes these methods attractive,  \citet{vulic-19-emnlp} recently show that even the most robust unsupervised method \cite{Artetxe-2018-acl} fails for a large number of language pairs. They suggest to rethink the main motivations behind fully unsupervised methods showing that with a small seed dictionary (500-1K pairs) their semi-supervised method always outperforms the unsupervised method and does not fail for any language pair. Other concurrent work \cite{ormazabal-19,doval-19} also advocates for weak supervision in {CLWE} methods.   


Almost all mapping-based {CLWE} methods, supervised and unsupervised alike, solve the  \emph{Procrustes} problem in the final step or during self-learning \cite{ruder-19-tutorial}. This restricts the transformation to be orthogonal linear mappings. However, learning an orthogonal linear mapping inherently assumes that the embedding spaces of different languages exhibit similar geometric structures (\ie\ approximately \emph{isomorphic}). Several recent studies have questioned this strong assumption and empirically showed that the {isomorphic} assumption does not hold in general even for two closely related languages like English and German \cite{Anders-18, patra-19}.

In this work, we propose \lnmap\ (\textbf{L}atent space \textbf{N}on-linear \textbf{Map}ping), a novel semi-supervised approach that uses \emph{non-linear} mapping in the latent space to learn CLWE. It uses minimal supervision from a seed dictionary, while leveraging semantic information from the monolingual word embeddings. As shown in Figure \ref{fig:proposed-model}, \lnmap\ comprises two \emph{autoencoders}, one for each language. The auto-encoders are first trained independently in a self-supervised way to induce the latent code space of the respective languages. Then, we use a small seed dictionary to learn the non-linear mappings between the two code spaces. To guide our mapping in the latent space, we include two additional constraints: back-translation and original embedding reconstruction. Crucially, our method does not enforce any strong prior constraints like the orthogonality (or isomorphic), rather it gives the model the flexibility to induce the required latent structures such that it is easier for the non-linear mappers to align them in the code space.

In order to demonstrate the effectiveness and robustness of \lnmap, we conduct extensive experiments on bilingual lexicon induction (BLI) with {fifteen (15)} different language pairs (in both directions) comprising high- and low-resource languages from two different datasets for different sizes of the seed dictionary. Our results show significant improvements for \lnmap\ over the state-of-the-art in most of the tested scenarios. It is particularly very effective for low-resource languages; for example, using 1K seed dictionary, \lnmap\ yields about 18\% absolute improvements on average over a state-of-the-art supervised method \cite{joulin-2018}. It also outperforms the most robust unsupervised system of \citet{Artetxe-2018-acl} in most of the translation tasks. Interestingly, for resource-rich language pairs, linear autoencoder performs better than non-linear ones.  
Our ablation study reveals the collaborative nature of \lnmap 's different components and efficacy of its non-linear mappings in the code space. We open-source our framework at \href{https://ntunlpsg.github.io/project/lnmap/}{https://ntunlpsg.github.io/project/lnmap/}.

\section{Background}
\label{sec:background}


\paragraph{Limitations of Isomorphic Assumption.}


Almost all {CLWE} methods inherently assume that embedding spaces of different languages are approximately \textit{isomorphic} (i.e., similar in geometric structure). However, recently researchers have questioned this simplified assumption and attributed the performance degradation of existing CLWE methods to {the} strong mismatches in embedding spaces caused by the linguistic and domain divergences \cite{Anders19CLWE,ormazabal-19}. \citet{Anders-18} empirically show that even closely related languages are far from being isomorphic. \citet{nakashole18} argue that mapping between embedding spaces of different languages can be approximately linear only at small local regions, but must be non-linear globally. \citet{patra-19} also recently show that etymologically distant language pairs cannot be aligned properly using orthogonal transformations.

\paragraph{Towards Semi-supervised Methods.}
A number of recent studies have questioned the robustness of existing unsupervised CLWE methods \cite{ruder-19-tutorial}. \citet{vulic-19-emnlp} show that even the most robust unsupervised method  \cite{Artetxe-2018-acl} fails for a large number of language pairs; it gives zero (or near zero) BLI performance for 87 out of 210 language pairs. With a seed dictionary of only 500 - 1000 word pairs, their supervised method outperforms unsupervised methods by a wide margin in most language pairs. Other recent work also suggested using semi-supervised methods \cite{patra-19, ormazabal-19}.

\paragraph{Mapping in Latent Space.}
\citet{mohi-joty:2019} propose adversarial autoencoder for \emph{unsupervised} word translation. They use \emph{linear} autoencoders in their model, and the mappers are also linear. They emphasize the benefit of using latent space over the original embedding space. Although their method is more robust than other existing adversarial models, still it suffers from training instability for distant language pairs. 

\paragraph{{Our Contributions.}} 
Our proposed \lnmap\ is independent of the isomorphic assumption. It uses weak supervision from a small seed dictionary, while leveraging rich structural information from monolingual embeddings. Unlike \citet{mohi-joty:2019}, the autoencoders in \lnmap\ are not limited to only linearity. More importantly, it uses \emph{non-linear} mappers. These two factors contribute to its robust performance even for very low-resource languages (\cref{sec:result}). To the best of our knowledge, we are the first to showcase such robust and improved performance with non-linear methods.\footnote{Our experiments with (unsupervised) adversarial training showed very unstable results with the non-linear mappers.}

\section{\lnmap\ Semi-supervised Framework}
\label{sec:model}

\begin{figure}[t!]
  \centering
\scalebox{1.}{
  \includegraphics[width=1\linewidth,trim=1 1 1 1,clip]{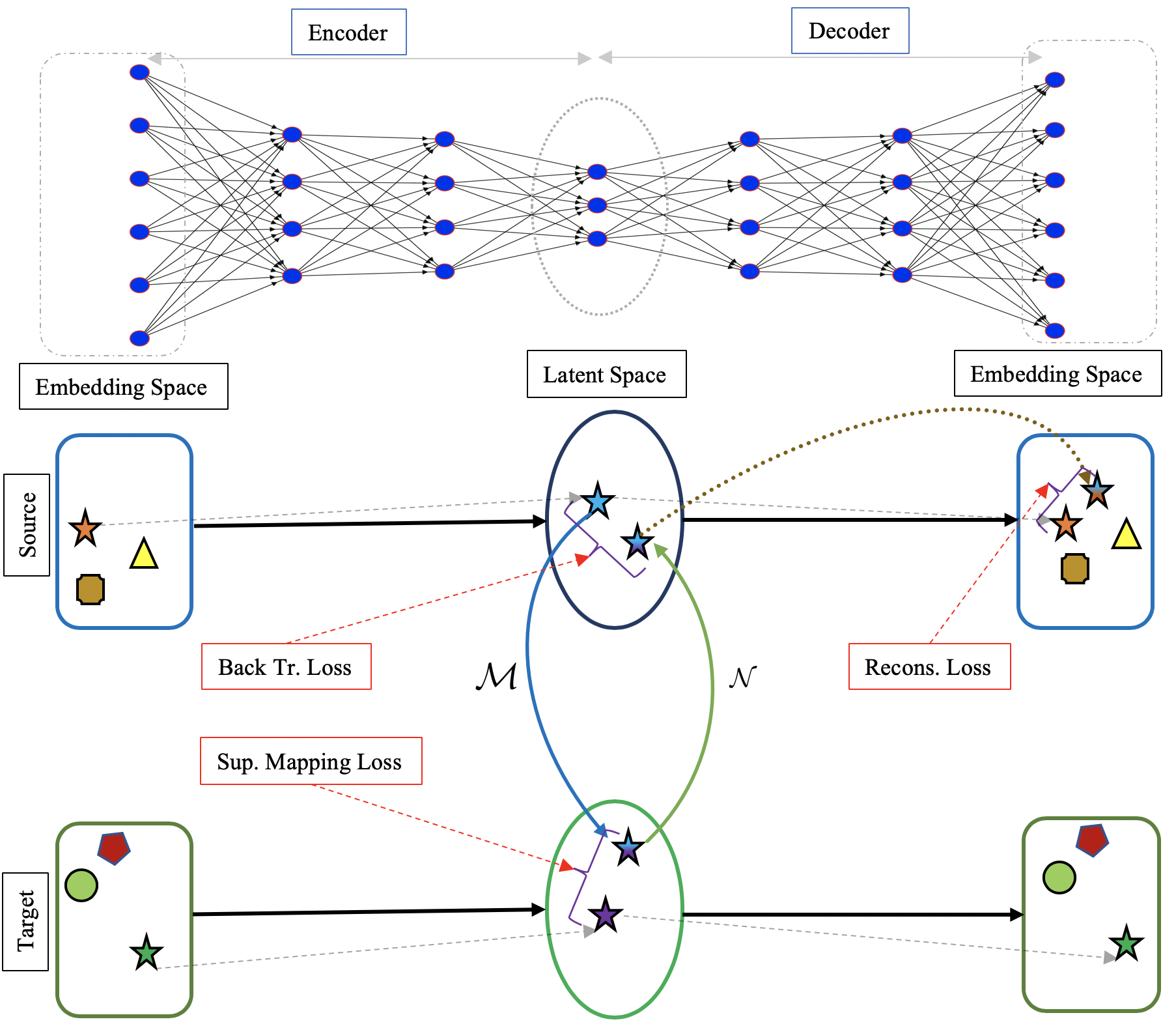}}
\caption{\lnmap: Our proposed semi-supervised framework.  Identical shapes with different colors denote the similar meaning words in different spaces (e.g., source/target embedding space or latent space).
\label{fig:proposed-model}}
\end{figure}

Let $\Vs_{\ell_x}$$=$$\{v_{x_1}, ..., v_{x_{n_x}}\}$ and $\Vs_{\ell_y}$$=$$\{v_{y_1}, ..., v_{y_{n_y}}\}$ be two sets of vocabulary consisting of $n_x$ and $n_y$ words for a source ($\ell_x$) and a target ($\ell_y$) language, respectively. Each word $v_{x_i}$ (resp. $v_{y_j}$) has an embedding $x_i \in \mathbb{R}^{d}$ (resp. $y_j \in \mathbb{R}^{d}$), trained with any word embedding models, \eg\ \texttt{FastText} \cite{bojanowski2017enriching}. Let $\Es_{\ell_x} \in \mathbb{R}^{n_x \times d}$ and $\Es_{\ell_y}  \in \mathbb{R}^{n_y \times d}$ be the word embedding matrices for the source and target languages, respectively. We are also given with a seed dictionary $\Ds$ $=$$\{(x_1, y_1), ..., (x_k, y_k)\}$ with $k$ word pairs. Our objective is to learn a transformation function $\Ms$ such that for any $v_{x_i} \in \Vs_{\ell_x}$, $\Ms(x_i)$ corresponds to its translation $y_j$, where $v_{y_j} \in \Vs_{\ell_y}$. Our approach \lnmap\ (Figure \ref{fig:proposed-model}) follows two sequential steps:
\begin{enumerate}[label=(\roman*)]
\vspace{-0.5em}
    \item Unsupervised latent space induction using monolingual autoencoders (\cref{subsec:unsup}), and \vspace{-0.5em}
    \item Supervised non-linear transformation learning with back-translation and source embedding reconstruction constraints (\cref{subsec:sup}).
\end{enumerate} 



\subsection{Unsupervised Latent Space Induction} \label{subsec:unsup}

We use two {autoencoders}, one for each language. Each autoencoder comprises an encoder $E_{\ell_x}$ (resp. $E_{\ell_y}$) and a decoder $D_{\ell_x}$ (resp. $D_{\ell_y}$). Unless otherwise stated, the autoencoders are \emph{non-linear}, where each of the encoder and decoder is a three-layer feed-forward neural network with two non-linear hidden layers. More formally, the encoding-decoding operations of the source autoencoder ($\texttt{autoenc}_{\ell_x}$) are defined as:

\begin{multicols}{2}
\noindent
\small
    \begin{eqnarray}
        h_1^{E_{\ell_x}} = && \hspace{-2em} \phi(\theta_1^{E_{\ell_x}} x_i) \\
        h_2^{E_{\ell_x}} = && \hspace{-2em} \phi(\theta_2^{E_{\ell_x}}h_1^{E_{\ell_x}})\\
        z_{x_i} = &&  \hspace{-2em} \theta_3^{E_{\ell_x}} h_2^{\ell_x} \label{eq:final_enc}
    \end{eqnarray}
    \begin{eqnarray}
       \hspace{-2em} h_1^{D_{\ell_x}} = && \hspace{-2em} \phi(\theta_3^{D_{\ell_x}}  z_{x_i})\\
      \hspace{-2em} h_2^{D_{\ell_x}}=  && \hspace{-2em} \phi(\theta_2^{D_{\ell_x}} h_1^{D_{\ell_x}})\\
      \hspace{-2em} \hat{x}_i = && \hspace{-2em} \phi(\theta_1^{D_{\ell_x}} h_2^{D_{\ell_x}}) \label{eq:final_dec}
    \end{eqnarray}
\end{multicols}
\normalsize

\noindent where $\theta_i^{E_{\ell_x}}$$\in$ $\mathbb{R}^{c_i \times d_i}$ and $\theta_i^{D_{\ell_x}}$$\in$ $\mathbb{R}^{d_i \times c_i}$ are the parameters of the layers in the encoder and decoder respectively, and $\phi$ is a non-linear activation function; we use Parametric  Rectified Linear Unit (\texttt{PReLU}) in all the hidden layers and \texttt{tanh} in the {final layer} of the decoder {(Eq. \ref{eq:final_dec})}. We use linear activations in the output layer of the encoder (Eq. \ref{eq:final_enc}). We train $\texttt{autoenc}_{\ell_x}$ with $l_2$ {reconstruction loss} as:

\begin{eqnarray}
\mathcal{L}_{\text{autoenc}_{\ell_x}}(\Theta_{E_{\ell_x}},\Theta_{D_{\ell_x}}) = \frac{1}{n_x} \sum_{i=1}^{n_x}  \| {x_i} - \hat{x}_i \|^2 \label{autoenc1loss} 
\end{eqnarray}
\normalsize

\noindent where $\Theta_{E_{\ell_x}}=\{\theta_1^{E_{\ell_x}}, \theta_2^{E_{\ell_x}}, \theta_3^{E_{\ell_x}} \}$ and $\Theta_{D_{\ell_x}}=\{\theta_1^{D_{\ell_x}}, \theta_2^{D_{\ell_x}}, \theta_3^{D_{\ell_x}} \}$ are the parameters of the encoder and the decoder of $\texttt{autoenc}_{\ell_x}$. 

The encoder, decoder and the reconstruction loss for the target autoencoder ($\texttt{autoenc}_{\ell_y}$) are similarly defined. 


\subsection{Supervised Non-linear Transformation}\label{subsec:sup}


Let $q(z_x|x)$ and $q(z_y|y)$ be the distributions of latent codes in $\texttt{autoenc}_{\ell_x}$ and $\texttt{autoenc}_{\ell_y}$, respectively. We have two non-linear mappers: $\Ms$ that translates a source code into a target code, and $\Ns$ that translates a target code into a source code (Figure \ref{fig:proposed-model}). Both mappers are implemented as a feed-forward neural network with a single hidden layer and \texttt{tanh} activations, and they are trained using the provided seed dictionary $\Ds$.

\paragraph{Non-linear Mapping Loss.} Let $\Theta_{\Ms}$ and $\Theta_{\Ns}$ denote the parameters of the two mappers $\Ms$ and $\Ns$, respectively. While mapping from $q(z_x|x)$ to $q(z_y|y)$, we jointly train the mapper $\Ms$ and the source encoder $E_{\ell_x}$ with the following $l_2$ loss.

\begin{eqnarray}
\mathcal{L}_{\text{MAP}}(\Theta_{\Ms},\Theta_{E_{\ell_x}}) = \frac{1}{k} \sum_{i=1}^{k}  \| {z_{y_i}} - \Ms(z_{x_i}) \|^2 \label{mappingloss} 
\end{eqnarray}
\normalsize

\noindent The mapping loss for $\Ns$ and  $E_{\ell_y}$ is similarity defined. To learn a better transformation function, we enforce two additional constraints to our objective -- back-translation and reconstruction.

\paragraph{Back-Translation Loss.} To ensure that a source code $z_{x_i} \in q(z_x|x)$ translated to the target language latent space $q(z_y|y)$, and then {translated back} to the original latent space remain unchanged, we enforce the back-translation constraint, that is,  $z_{x_i} \rightarrow \Ms(z_{x_i}) \rightarrow \Ns(\Ms(z_{x_i})) \approx z_{x_i}$. The back-translation (BT) loss from $q(z_y|y)$ to $q(z_x|x)$ is  

\begin{eqnarray}
\hspace{-3em} 
\mathcal{L}_{\text{BT}}(\Theta_{\Ms},\Theta_{\Ns})= && \nonumber \\ && \hspace{-4em}  \frac{1}{k} \sum_{i=1}^{k}  \| {z_{x_i}} -  \Ns({\Ms({z_{x_i}}})) \|^2 \label{backtrloss}
\end{eqnarray}
\normalsize

\noindent The BT loss in the other direction $(z_{y_j}$$\rightarrow$$\Ns(z_{y_j})$$\rightarrow$ $\Ms(\Ns(z_{y_j})) \approx z_{y_j})$ is similarly defined.

\paragraph{Reconstruction Loss.}  In addition to  back-translation, we include another constraint to guide the mapping further. In particular, we ask the decoder $D_{\ell_x}$ of $\texttt{autoenc}_{\ell_x}$ to reconstruct the original embedding $x_i$ from the back-translated code {$\Ns(\Ms({z_{x_i}}))$}. We compute this {original embedding reconstruction loss} for $\texttt{autoenc}_{\ell_x}$ as:

\begin{eqnarray}
 \hspace{-2em}  \mathcal{L}_{\text{REC}}(\theta_{E_{\ell_x}}, \theta_{D_{\ell_x}}, \Theta_{\Ms}, \Theta_{\Ns}) = && \nonumber \\  && \hspace{-10em} \frac{1}{k} \sum_{i=1}^{k}  \|  {x_i} - D_{\ell_x}(\Ns(\Ms{z_{x_i}}))) \|^2 
 \label{reconsAloss}
\end{eqnarray}
\normalsize

\noindent The reconstruction loss for $\texttt{autoenc}_{\ell_y}$ is defined similarly. Both back-translation and reconstruction lead to more \textit{stable training} in our experiments. In our ablation study (\cref{subsec:dissection}), we empirically show the efficacy of the addition of these two constraints. 


\paragraph{Total Loss.} The total loss for mapping a batch of word embeddings from source to target is:
\begin{equation}
\mathcal{L}_{{\ell_x} \rightarrow {\ell_y}} =  \mathcal{L}_{\text{MAP}} + \lambda_1  \mathcal{L}_{\text{BT}} + \lambda_2 \mathcal{L}_{\text{REC}} \label{eq:totalloss}
\end{equation}
\normalsize

\noindent where $\lambda_1$ and $\lambda_2$ control the relative importance of the loss components. Similarly we define the total loss for mapping in the opposite direction $\mathcal{L}_{{\ell_y} \rightarrow {\ell_x}}$.


\paragraph{Remark.} Note that our approach is fundamentally different from existing methods  in two ways. First, most of the existing methods directly map the distribution of the source embeddings $p(x)$ to the distribution of the  {target embeddings $p(y)$}. Second, they learn a linear mapping function assuming that the two languages' embedding spaces are nearly isomorphic, which does not hold in general  \cite{Anders-18,patra-19}.

Mapping the representations in the code space using non-linear transformations gives our model the flexibility to induce the required semantic structures in its {latent space} that could potentially yield more accurate cross-lingual mappings (\cref{sec:result}).

\subsection{Training Procedure}

We present the training method of \lnmap\ in Algorithm \ref{alg:training}.
In the first step, we  pre-train $\texttt{autoenc}_{\ell_x}$ and $\texttt{autoenc}_{\ell_y}$ separately on the respective monolingual word embeddings. In this unsupervised step, we use the first 200K embeddings. \red{This pre-training induce word semantics (and relations) in the code space \cite{mohi-joty:2019}.}

\SetKwRepeat{Do}{do}{while}
\begin{algorithm2e*}[t!]
\SetKwInOut{Input}{Input}\SetKwInOut{Output}{Output}
\SetAlgoLined
\SetNoFillComment
\LinesNotNumbered 
\SetNlSkip{0em}
\Input{Word embedding matrices: $\Es_{\ell_x}$, $\Es_{\ell_y}$, seed dictionary: $\Ds$, and {increment count} $C$}
\tcp{Unsup. latent space induction}
1.  Train $\texttt{autoenc}_{\ell_x}$ and $\texttt{autoenc}_{\ell_y}$ separately for some epochs on monolingual word embeddings  \\
\tcp{Sup. non-linear transformation}
2. $iter = 0;  \Ds_{\text{orig}} = \Ds$ \\

3. \Do{not converge}{
    \hspace{-1em} $iter = iter + 1$ \\
    \hspace{-1em} \Ni \For{n\_epochs}{ 
        \hspace{-1em} \Na Sample a mini-batch from $\Ds$  \\
        \hspace{-1em} \Nb Update mapper ${\Ms}$ and $E_{\ell_x}$ on the non-linear mapping loss \\
        \hspace{-1em} \Nc Update mappers $\Ms$ and $\Ns$ on the  back-translation loss \\
        \hspace{-1em} \Nd Update mappers ($\Ms$, $\Ns$) and $\texttt{autoenc}_{\ell_x}$ on the reconstruction loss  \\
        }
    \hspace{-1em} \Nii Induce a new dictionary $\Ds_{\text{new}}$ of size: {\textit{$iter \times C$}}\\
    \hspace{-1em} \Niii Create a new dictionary, $\Ds = \Ds_{\text{orig}} \bigcup  \Ds_{\text{new}} $
}

\caption{Training \lnmap}
\label{alg:training}
\end{algorithm2e*}
\normalsize

The next step is the self-training process, where we train the mappers along with the autoencoders using the seed dictionary in an iterative manner. We keep a copy of the original dictionary $\Ds$; let us call it $\Ds_{\text{orig}}$. We first update the mapper $\Ms$ and the source encoder $E_{\ell_x}$ on the mapping loss (Eq. \ref{mappingloss}). The mappers (both $\Ms$ and $\Ns$) then go through two more updates, one for back-translation (Eq. \ref{backtrloss}) and the other for reconstruction of the source embedding (Eq. \ref{reconsAloss}). The entire source autoencoder $\texttt{autoenc}_{\ell_x}$ (both $E_{\ell_x}$ and $D_{\ell_x}$) in this stage gets updated only on the reconstruction loss. 


After each iteration of training (step \emph{i.} in Alg. \ref{alg:training}), we induce a new dictionary $\Ds_{\text{new}}$ using the learned encoders and mappers. To find the nearest target word ($y_j$) of a  source word ($x_i$) in the target latent space, we use the Cross-domain Similarity Local Scaling ({CSLS}) measure which works better than simple cosine similarity in mitigating the \textit{hubness} problem \cite{conneau2018word}. It penalizes the words that are close to many other words in the target latent space. To induce the dictionary, we compute {CSLS} for $K$ 
most frequent source and target words 
and select the translation pairs that are nearest neighbors of each other according to {CSLS}.

For the next iteration of training, we construct the dictionary $\Ds$ by merging $\Ds_{\text{orig}}$ with the $l$ most similar (based on CSLS) word pairs from $\Ds_{\text{new}}$. We set $l$ as $l = {iter} \times C$, where $iter$ is the current iteration number and $C$ is a hyperparameter. This means we incrementally update the dictionary size. This is because the induced dictionary at the initial iterations is likely to be noisy. As the training progresses, the model becomes more mature, and the induced dictionary pairs become better. For convergence, we use the criterion: if the difference between the average similarity scores of two successive iteration steps is less than a threshold (we use $1e^{-6}$), then stop the training process.






\section{Experimental Settings}
\label{sec:setting}



We evaluate our approach on bilingual lexicon induction, also known as \emph{word translation}.

\subsection{Datasets}
To demonstrate the effectiveness of our method, we evaluate our models against baselines on two popularly used datasets: {MUSE} \cite{conneau2018word} and  {VecMap} \cite{Dinu-iclr-workshop15}. 

The MUSE dataset consists of FastText monolingual embeddings of 300 dimensions \cite{bojanowski2017enriching} trained on Wikipedia monolingual corpus and gold dictionaries for 110 language pairs.\footnote{\href{https://github.com/facebookresearch/MUSE}{https://github.com/facebookresearch/MUSE}} 
{To show the generality of different methods, we consider $15$ different language pairs with $15\times 2=30$ different translation tasks} encompassing resource-rich and low-resource languages from different language families. {In particular, we evaluate on English (En) from/to Spanish (Es), German (De), Italian (It), Russian (Ru), Arabic (Ar), Malay (Ms), Finnish (Fi), Estonian (Et), Turkish (Tr), Greek (El), Persian (Fa), Hebrew (He), Tamil (Ta), Bengali (Bn), and Hindi (Hi). We differentiate between high- and low-resource languages by the availability of NLP-resources in general.}




{The VecMap dataset \cite{Dinu-iclr-workshop15, artetxe2018aaai} is a more challenging dataset and contains monolingual embeddings for English, Spanish, German, Italian, and Finnish.\footnote{\href{https://github.com/artetxem/vecmap/}{https://github.com/artetxem/vecmap/}} According to \newcite{Artetxe-2018-acl}, existing unsupervised methods often fail to produce meaningful results on this dataset. English, Italian, and German embeddings were trained on WacKy crawling corpora using CBOW \cite{Mikolov-word2vec}, while Spanish and Finnish embeddings were trained on WMT News Crawl and Common Crawl, respectively. 
} 




\subsection{Baseline Methods}
We compare our proposed \lnmap\ with several existing methods comprising {supervised, semi-supervised, and unsupervised models.} For each baseline model, we conduct experiments with the publicly available code. In the following, we give a brief description of the baseline models.

\paragraph{Supervised {\& Semi-supervised} Methods.}

\begin{description}[leftmargin=0pt,itemsep=-0.0em]
\item [\Na \newcite{artetxe2017acl} ] propose a \textit{self-learning framework} that performs two steps iteratively until convergence. In the first step, they use the dictionary (starting with the seed dictionary) to learn a linear mapping, which is then used in the second step to induce a new dictionary. 
\item [\Nb \newcite{artetxe2018aaai} ] propose a \textit{multi-step framework} that generalizes previous studies. Their framework consists of several steps: whitening, orthogonal mapping, re-weighting, de-whitening, and dimensionality reduction.
\item [\Nc \newcite{conneau2018word} ] compare their unsupervised model with a supervised baseline that learns an orthogonal mapping between the embedding spaces by iterative Procrustes refinement. They also propose CSLS for nearest neighbour search. 
\item[\Nd \newcite{joulin-2018} ] show that minimizing a convex relaxation of the CSLS loss significantly improves the quality of bilingual word vector alignment. Their method achieves state-of-the-art results for many languages \cite{patra-19}.
\item[\Ne \newcite{jawanpuria2018learning}] {propose a geometric approach where they decouple CLWE learning into two steps: (i) learning rotations for language-specific embeddings to align them to a common space, and (ii) learning a similarity metric in the common space to model similarities between the embeddings \blue{of the two languages}.}
\item[\Nf \newcite{patra-19}] {propose a semi-supervised technique that relaxes the isomorphic assumption while leveraging both seed dictionary pairs and a larger set of unaligned word embeddings.}

\end{description}

\paragraph{Unsupervised Methods.}
\begin{description}[leftmargin=0pt,itemsep=-0.0em]

\item [\Na \newcite{conneau2018word}] are the first to show impressive results for unsupervised word translation by pairing adversarial training with effective refinement methods. Given two monolingual word embeddings, their adversarial training plays a \textit{two-player game}, where a linear mapper (generator) plays against a discriminator. They also impose the orthogonality constraint on the mapper. After adversarial training, they use the iterative Procrustes solution similar to their supervised approach.   

\item [\Nb \newcite{Artetxe-2018-acl}] learn an initial dictionary by exploiting the structural similarity of the embeddings in an unsupervised way. They propose a robust self-learning to improve it iteratively. This model is by far the most robust and best performing unsupervised model \cite{vulic-19-emnlp}.

\item [\Nc \newcite{mohi-joty:2019}] use adversarial autoencoder for unsupervised word translation. They use linear autoencoders in their model, and the mappers are also linear. 
\end{description}


\subsection{Model Variants and Settings}
{We experiment with two variants of our model:  the default \lnmap\  that uses non-linear autoencoders and \lnmaplin\ that uses linear autoencoders. In both the variants, the mappers are non-linear.}
{We train our models using stochastic gradient descent ({SGD}) with a batch size of 128, a learning rate of $1e^{-4}$, and a step learning rate decay schedule. During the dictionary induction process in each iteration, we consider \blue{$K=15000$} most frequent words from the source and target languages. For dictionary update, we set $C=2000$.}

\section{Results and Analysis}
\label{sec:result}

We present our results on low-resource and resource-rich languages from MUSE dataset in Tables \ref{tab:low-resource-results} and \ref{tab:high-resource-results}, respectively, and the results on VecMap dataset in Table \ref{tab:dinu-results}. We present the results in \textit{precision@1}, which means how many times one of the correct translations of a source word is predicted as the top choice. For each of the cases, we show results on seed dictionary of three different sizes including 1-to-1 and 1-to-many mappings;  \textbf{``1K Unique''} and \textbf{``5K Unique''} contain 1-to-1 mappings of 1000 and 5000 source-target pairs respectively, while \textbf{``5K All''} contains 1-to-many mappings of all 5000 source and target words, that is, for each source word there can be multiple target words. Through experiments and analysis, our goal is to assess the following questions. 
\begin{enumerate}[label=(\roman{*}),topsep=0.5em]
    \item Does \lnmap\ improve over the best existing methods in terms of mapping accuracy on low-resource languages (\cref{subsec:low-resource-con})?
    \item How well does \lnmap\ perform on resource-rich languages (\cref{subsec:high-resource-con-dinu})?
    \item \blue{What is the effect of non-linearity in the autoencoders?} (\cref{subsec:lin-ae})
    \item Which components of \lnmap\ attribute to improvements (\cref{subsec:dissection})?
\end{enumerate}



\subsection{Performance on Low-resource Languages} \label{subsec:low-resource-con}


\begingroup
\setlength{\tabcolsep}{3pt}
\begin{table*}[t!]
\centering
\footnotesize
\scalebox{0.76}{\begin{tabular}{l|cc|cc|cc|cc|cc|cc|cc|cc|cc|cc||c}

&\multicolumn{2}{c}{\textbf{En-Ms}} & \multicolumn{2}{c}{\textbf{En-Fi}} &    \multicolumn{2}{c}{\textbf{En-Et}} & \multicolumn{2}{c}{\textbf{En-Tr}} & \multicolumn{2}{c}{\textbf{En-El}} & \multicolumn{2}{c}{\textbf{En-Fa}} & \multicolumn{2}{c}{\textbf{En-He}} & \multicolumn{2}{c}{\textbf{En-Ta}} & \multicolumn{2}{c}{\textbf{En-Bn}} & \multicolumn{2}{c}{\textbf{En-Hi}} &
\textbf{\blue{Avg.}}
\\

& \textbf{$\rightarrow$} & \textbf{$\leftarrow$} & 
  \textbf{$\rightarrow$} & \textbf{$\leftarrow$} & 
  \textbf{$\rightarrow$} & \textbf{$\leftarrow$} &
  \textbf{$\rightarrow$} & \textbf{$\leftarrow$} & 
  \textbf{$\rightarrow$} & \textbf{$\leftarrow$} & 
  \textbf{$\rightarrow$} & \textbf{$\leftarrow$} &
  \textbf{$\rightarrow$} & \textbf{$\leftarrow$} & 
  \textbf{$\rightarrow$} & \textbf{$\leftarrow$} & 
  \textbf{$\rightarrow$} & \textbf{$\leftarrow$} &
  \textbf{$\rightarrow$} & \textbf{$\leftarrow$} &
\\ 
\midrule
\textbf{GH Distance} & 
\multicolumn{2}{c|}{0.49} & \multicolumn{2}{c|}{0.54} & \multicolumn{2}{c|}{0.68} & \multicolumn{2}{c|}{0.41} & \multicolumn{2}{c|}{0.46} & \multicolumn{2}{c|}{0.39} & \multicolumn{2}{c|}{0.45} & \multicolumn{2}{c|}{0.47} & \multicolumn{2}{c|}{0.49} & 
\multicolumn{2}{c||}{0.56}  &        
\\ 
\midrule
\multicolumn{11}{l}{\textbf{Unsupervised Baselines}}  \\

\newcite{Artetxe-2018-acl} & 49.0 & 49.7 & 49.8 & 63.5 & 33.7 & 51.2 & 52.7 & 63.5 & 47.6 & 63.4 & 33.4 & 40.7 & 43.8 & 57.5 & 0.0 & 0.0 & 18.4 & 23.9 & 39.7 & 48.0 & \blue{41.5} \\

\newcite{conneau2018word} &  46.2 & 0.0 & 38.4 & 0.0 & 19.4 & 0.0 & 46.4 & 0.0 & 39.5 & 0.0 & 30.5 & 0.0 & 36.8 & 53.1 & 0.0 & 0.0 & 0.0 & 0.0 & 0.0 & 0.0 & \blue{15.5} \\

\newcite{mohi-joty:2019}  & \blue{54.1} & \blue{51.7} &  \blue{44.8} &  \blue{62.5} & \blue{31.8} & \blue{48.8} & \blue{51.3} & \blue{61.7} & \blue{47.9} & \blue{63.5} & \blue{36.7} & \blue{44.5} & \blue{44.0} & \blue{57.1} & \blue{0.0} & \blue{0.0} & \blue{0.0} & \blue{0.0} & \blue{0.0} & \blue{0.0} & \blue{35.0}\\

\midrule
\multicolumn{20}{c}{{\small{{Supervision With \hl{\textbf{``1K Unique''}} Seed Dictionary}}}} \\
\midrule

\multicolumn{11}{l}{\textbf{Sup.\blue{/Semi-sup.} Baselines}}  \\

\newcite{artetxe2017acl} & 36.5 & 41.0 & 40.8 & 56.0 & 21.3 & 39.0 & 39.5 & 56.5 & 34.5 & 56.2 & 24.1 & 35.7 & 30.2 & 51.7 & 5.4 & 12.7 & 6.2 & 19.9 & 22.6 & 38.8 & \blue{33.5}\\

\newcite{artetxe2018aaai} & 35.3 & 34.0 & 30.8 & 40.8 & 21.6 & 32.6 & 33.7 & 43.3 & 32.0 & 46.4 & 22.8 & 27.6 & 32.27 & 39.1 & 7.3 & 11.9 & 11.3 & 15.7 & 26.2 & 30.7  & \blue{28.8}\\

\newcite{conneau2018word} & 46.2 & 44.7 & 46.0 & 58.4 & 29.3 & 40.0 & 44.8 & 58.5 & 42.1 & 56.5 & 31.6 & 38.4 & 38.3 & 52.4 & 11.7 & 16.0 & 14.3 & 19.7 & 32.5 & 42.3  & \blue{38.2}\\

 \newcite{joulin-2018} & 31.4 & 30.7 & 30.4 & 41.4 & 20.1 & 26.0 & 30.7 & 36.5 & 28.8 & 43.6 & 18.7 & 23.1 & 33.5 & 34.3 & 6.0 & 10.1 & 7.6 & 11.3 & 20.7 & 25.7  & \blue{25.6}\\

\newcite{jawanpuria2018learning} & \blue{40.0} & \blue{39.6} & \blue{37.5} & \blue{50.7} & \blue{24.9} & \blue{38.4} & \blue{39.7} & \blue{49.7} & \blue{36.6} & \blue{52.9} & \blue{26.1} & \blue{33.0} & \blue{35.1} & \blue{44.5} & \blue{10.0} & \blue{15.9} & \blue{12.0} & \blue{19.7} & \blue{30.5} & \blue{37.1}  & \blue{33.7}  \\

\newcite{patra-19}  & \blue{40.4} & \blue{41.4} & \blue{44.3} & \blue{59.8} & \blue{21.0} & \blue{40.4} & \blue{41.4} & \blue{58.8} & \blue{37.1} & \blue{58.9} & \blue{26.5} & \blue{39.6} & \blue{38.4} & \blue{54.1} & \blue{6.4} & \blue{15.1} & \blue{6.1} & \blue{18.1} & \blue{24.9} & \blue{35.4}   & \blue{35.4}\\
 

\lnmapb & \textbf{50.6} & \textbf{49.5} & \textbf{52.5} & \textbf{62.1} & \textbf{38.2} & \textbf{49.4} & \textbf{52.6} & \textbf{62.1} & \textbf{48.2} & \textbf{58.9} & \textbf{35.5} & \textbf{40.9} & \textbf{46.6} & \textbf{52.8} & \textbf{17.6} & \textbf{21.2} & \textbf{18.4} & \textbf{27.2} & \textbf{37.1} & \textbf{47.4} & \textbf{43.4}
\\

\lnmaplin & 49.8 &  48.7 &  48.5 &  61.2 &  36.5 &  49.1 &  49.3 &  61.9 &  47.2 &  58.3 &  34.7 &  40.1 &  43.0 &  52.3 &  14.5 &  20.3 &  16.5 &  26.1 &  35.6 &  46.6 & 42.1

\\
\midrule
\multicolumn{20}{c}{{\small{Supervision With \hl{\textbf{``5K Unique''}} Seed Dictionary}}} \\
\midrule

\multicolumn{11}{l}{\textbf{Sup.\blue{/Semi-sup.} Baselines}}  \\

\newcite{artetxe2017acl} &36.5 & 42.0 & 40.8 & 57.0 & 22.4 & 39.6 & 39.6 & 56.7 & 37.2 & 56.4 & 26.0 & 35.3 & 31.6 & 51.9 & 6.2 & 13.4 & 8.2 & 21.3 & 23.2 & 38.3   & \blue{34.2} \\

\newcite{artetxe2018aaai} & 54.6 & 52.5 & 48.8 & 65.2 & 38.2 & 54.8 & 52.0 & 65.1 & 47.5 & 64.6 & 38.4 & 42.4 & 47.4 & \textbf{57.4} & 18.4 & 25.8 & 21.9 & 31.8 & \textbf{40.3} & 49.5  & \blue{45.8}\\

\newcite{conneau2018word} & 46.4 & 45.7 & 46.0 & 59.2 & 31.0 & 41.7 & 45.9 & 60.1 & 43.1 & 56.8 & 31.6 & 37.7 & 38.4 & 53.4 & 14.3 & 19.1 & 15.0 & 22.6 & 32.9 & 42.8  & \blue{39.2}\\

\newcite{joulin-2018} & 50.0 & 49.3 & \textbf{53.0} & 66.1 & 39.8 & 52.0 & \textbf{54.0} & 61.7 & 47.6 & 63.4 & \textbf{39.6} & 42.2 & \textbf{53.0} & 56.3 & 16.0 & 24.2 & 21.3 & 27.0 & 38.3 & 47.5  & \blue{45.2}\\

\newcite{jawanpuria2018learning} & {\blue{51.0}} & \blue{49.8} & \blue{47.4} & \blue{65.1} & \blue{36.0} & \blue{49.8} & \blue{49.3} & \blue{63.9} & \blue{46.6} & \blue{62.3} & \blue{36.6} & \blue{40.8} & \blue{44.1} & \blue{56.1} & \blue{16.1} & \blue{23.2} & \blue{18.6} & \blue{25.9} & \blue{37.5} & \blue{45.9}   & \blue{43.3} \\

\newcite{patra-19}   & \blue{46.0} & \blue{46.7} & \blue{48.6} & \blue{60.9} & \blue{33.1} & \blue{47.2} & \blue{48.3} & \blue{61.0} & \blue{44.2} & \blue{60..9} & \blue{34.4} & \blue{40.7} & \blue{43.5} & \blue{56.5} & \blue{15.3} & \blue{22.0} & \blue{15.2} & \blue{25.0} & \blue{34.7} & \blue{43.5}   & \blue{41.4}\\
 

\lnmapb & \textbf{51.3} & \textbf{54.2} & {52.7} & \textbf{67.9} & \textbf{40.2} & \textbf{56.4} & {53.1} & \textbf{65.5} & \textbf{48.2} & \textbf{64.8} & {36.2} & \textbf{44.4} & {47.5} & {56.6} & \textbf{19.7} & \textbf{31.5} & \textbf{22.0} & \textbf{36.2} & {38.5} & \textbf{52.2} & \textbf{46.9}
\\

\lnmaplin & 50.1 &  53.9 &  51.3 &  67.0 &  38.6 &  55.6 &  51.1 &  64.9 &  47.7 &  63.6 &  35.6 &  44.0 &  44.2 &  55.9 &  18.6 &  27.3 &  19.6 &  31.6 &  36.5 &  51.3 & 45.4
 
\\
\midrule
\multicolumn{20}{c}{{\small{Supervision With \hl{\textbf{``5K All''}} (``5K Unique'' Source Words) Seed Dictionary}}} \\
\midrule

\multicolumn{11}{l}{\textbf{Sup.\blue{/Semi-sup.} Baselines}}  \\

\newcite{artetxe2017acl} & 37.0 & 41.6 & 40.8 & 57.0 & 22.7 & 39.5 & 38.8 & 56.9 & 37.5 & 57.2 & 25.4 & 36.3 & 32.2 & 52.1 & 5.9 & 14.1 & 7.7 & 21.7 & 22.4 & 38.3  & \blue{34.3}\\

\newcite{artetxe2018aaai} & 55.2 & 51.7 & 48.9 & 64.6 & 37.4 & 54.0 & 52.2 & 63.7 & 48.2 & 65.0 & 39.0 & 42.6 & 47.6 & 58.0 & \textbf{19.6} & 25.2 & 21.1 & 30.6 & \textbf{40.4} & 50.0   & \blue{45.8}\\

\newcite{conneau2018word} & 46.3 & 44.8 & 46.4 & 59.0 & 30.9 & 42.0 & 45.8 & 59.0 & 44.4 & 57.4 & 31.8 & 38.8 & 39.0 & 53.4 & 15.1 & 18.4 & 15.5 & 22.4 & 32.9 & 44.4  & \blue{39.4}\\

\newcite{joulin-2018} & \textbf{51.4} & 49.1 & \textbf{55.6} & 65.8 & 40.0 & 50.2 & \textbf{53.8} & 61.7 & \textbf{49.1} & 62.8 & \textbf{40.5} & 42.4 & \textbf{52.2} & 57.9 & 17.7 & 24.0 & 20.2 & 26.9 & 38.2 & 47.1  & \blue{45.3}\\

\newcite{jawanpuria2018learning}  & \textbf{\blue{51.4}} & \blue{47.7} & \blue{46.7} & \blue{63.4} & \blue{33.7} & \blue{48.7} & \blue{48.6} & \blue{61.9} & \blue{46.3} & \blue{61.8} & \blue{38.0} & \blue{40.9} & \blue{43.1} & \blue{56.7} & \blue{16.5} & \blue{23.1} & \blue{19.3} & \blue{25.6} & \blue{37.7} & \blue{44.1}   & \blue{42.8}\\

\newcite{patra-19}   & \blue{48.4} & \blue{43.8} & \blue{53.2} & \blue{63.8} & \blue{36.3} & \blue{48.3} & \blue{51.8} & \blue{59.6} & \blue{48.2} & \blue{61.8} & \blue{38.4} & \blue{39.3} & \blue{51.6} & \blue{55.2} & \blue{16.5} & \blue{22.7} & \blue{17.5} & \blue{26.7} & \blue{36.2} & \blue{45.4}   & \blue{43.3}\\
 

\lnmapb &  {50.3} & \textbf{54.1} & {53.1} & \textbf{70.5} & \textbf{41.2} & \textbf{57.5} & {52.5} & \textbf{65.3} & \textbf{49.1} & \textbf{66.6} & {36.8} & \textbf{43.7} & {47.6} & \textbf{59.2} & {18.9} & \textbf{32.1} & \textbf{21.4} & \textbf{35.2} & {37.6} & \textbf{51.6} & \textbf{47.2}
\\

\lnmaplin & 50.0 &  53.2 &  51.2 &  67.5 &  39.9 &  54.5 &  50.9 &  64.2 &  48.6 &  66.1 &  36.4 &  42.9 &  44.6 &  59.0 &  18.0 &  28.7 &  20.1 &  30.8 &  37.1 &  50.5 & 46.7

\\
\bottomrule
\end{tabular}}
\caption{Word translation accuracy (P@1) on \textbf{low-resource} languages on \textbf{MUSE dataset} using {fastText}. } 
\label{tab:low-resource-results} 
\end{table*}
\endgroup

Most of the unsupervised models fail in the majority of the low-resource languages \cite{vulic-19-emnlp}. On the other hand, the performance of supervised models on low-resource languages was not satisfactory, \blue{especially with small seed dictionary.} Hence, we first compare \lnmap 's performance on these languages. {From Table \ref{tab:low-resource-results},  we see that on average \lnmap\ outperforms every baseline by a good margin \blue{(1.1\% - 5.2\% from the best baselines)}.}

For {``1K Unique''} dictionary, \lnmap\ exhibits impressive performance. In all the 20 translation tasks, it outperforms all the (semi-)supervised baselines by a wide margin. If we compare with \citet{joulin-2018}, a state-of-the-art supervised model, \lnmap's average improvement is $\sim$$18\%$, which is remarkable. Compared to other baselines, the average margin of improvement is also quite high -- $9.9\%, 14.6\%, 5.2\%, 9.7\%,$ and $8.0\%$ gains over \citet{artetxe2017acl}, \newcite{artetxe2018aaai}, \newcite{conneau2018word}, \newcite{jawanpuria2018learning}, and \newcite{patra-19}, respectively. We see that among the supervised baselines, \citet{conneau2018word}'s model performs better than others.


If we increase the dictionary size, we can still see the dominance of \lnmap\ over the baselines. For {{``5K Unique''}} seed dictionary, it performs better than the baselines on {14/20 translation tasks}, while for {``5K All''} seed dictionary, the best performance by \lnmap\ is on {13/20 translation tasks.} 

One interesting thing to observe is that, under resource-constrained setup \lnmap's performance is {impressive}, making it suitable for very low-resource languages like En-Ta, En-Bn, and En-Hi. 


Now if we look at the performance of unsupervised baselines on low-resource languages, we see that \citet{conneau2018word}'s model fails to converge on the majority of the translation tasks (12/20), {while the model of \citet{mohi-joty:2019} fails to converge on En$\leftrightarrow$Ta, En$\leftrightarrow$Bn, and En$\leftrightarrow$Hi.} Although the most robust unsupervised method of \citet{Artetxe-2018-acl} performs better than the other unsupervised approaches, it still fails to converge on En$\leftrightarrow$Ta tasks. If we compare its performance with \lnmap, we see that our model outperforms the best unsupervised model of \citet{Artetxe-2018-acl} on 18/20 low-resource translation tasks.


\begingroup
\begin{table*}[t!]
\centering
\small
\scalebox{0.9}{\begin{tabular}{l|cc|cc|cc|cc|cc||c}
&\multicolumn{2}{c}{\textbf{En-Es}}  &        \multicolumn{2}{c}{\textbf{En-De}} &\multicolumn{2}{c}{\textbf{En-It}}& \multicolumn{2}{c}{\textbf{En-Ar}} &\multicolumn{2}{c}{\textbf{En-Ru}}&
\textbf{\blue{Avg.}}
\\
& \textbf{$\rightarrow$} & \textbf{$\leftarrow$} & \textbf{$\rightarrow$} & \textbf{$\leftarrow$} & 
\textbf{$\rightarrow$} & \textbf{$\leftarrow$} & 
\textbf{$\rightarrow$} & \textbf{$\leftarrow$} & 
\textbf{$\rightarrow$} & \textbf{$\leftarrow$} 
\\   
\midrule
\textbf{GH Distance} & \multicolumn{2}{c|}{0.21} & \multicolumn{2}{c|}{0.31} & \multicolumn{2}{c|}{0.19} & \multicolumn{2}{c|}{0.46} & \multicolumn{2}{c}{0.46} &

\\
\midrule

\multicolumn{6}{l}{\textbf{Unsupervised Baselines}} & \\

\newcite{Artetxe-2018-acl} & 82.2 & 84.4 & 74.9 & 74.1 & 78.9 & 79.5 & 33.2 & 52.8 & 48.93 & 65.0 & 67.4 \\

\newcite{conneau2018word} & 81.8 & 83.7 & 74.2 & 72.6 & 78.3 & 78.1 & 29.3 & 47.6 & 41.9 & 59.0 & 64.7  \\

\newcite{mohi-joty:2019} & \blue{82.7} & \blue{84.7} & \blue{75.4} & \blue{74.3} & \blue{79.0} & \blue{79.6} & \blue{36.3} & \blue{52.6} & \blue{46.9} & \blue{64.7} & 67.6 \\

\midrule
\multicolumn{10}{c}{{\small{Supervision With \hl{\textbf{``1K Unique''}} Seed Dictionary}}} \\
\midrule

\multicolumn{6}{l}{\textbf{Sup.\blue{/Semi-sup.} Baselines}}  \\

\newcite{artetxe2017acl} &  81.0 & {83.6} & 73.8 & 72.4 & 76.6 & {77.8} & 24.9 & 44.9 & 46.3 & 61.7 & 64.3 \\

\newcite{artetxe2018aaai} & 73.8 & 76.6 & 62.5 & 57.6 & 67.9 & 70.0 & 25.8 & 37.3 & 40.2 & 49.5 & 56.2\\

\newcite{conneau2018word} & {81.2} & 82.8 & 73.6 & 73.0 & {77.6} & 76.6 & 34.7 & 46.4 & 48.5 & 60.6 & 65.5\\

\newcite{joulin-2018} & 70.8 & 74.1 & 59.0 & 54.0 & 62.7 & 67.2 & 22.4 & 32.2 & 39.6 & 45.4 & 52.8\\
 
\newcite{jawanpuria2018learning}  & \blue{75.1} & \blue{77.3} & \blue{66.0} & \blue{62.6} & \blue{69.3} & \blue{71.6} & \blue{28.4} & \blue{40.6} & \blue{41.7} & \blue{53.9}  & 58.6\\
 
\newcite{patra-19}  & {\blue{81.9}} & {\blue{83.8}} & \blue{74.6} & {{73.1}} & {{\blue{78.0}}} & {\blue{78.1}} & \blue{29.8} & \blue{50.9} & \blue{46.3} & {63.6} & 66.0\\ 
 


\lnmap & 80.1 &  80.2 &  73.3 &  71.8 &  77.1 &  75.2 &  \textbf{40.5} &  52.2 &  49.9 &  62.1 & 66.2
\\

\lnmaplinb & \textbf{83.2} &  \textbf{85.5} &  \textbf{76.2} &  \textbf{74.9} &  \textbf{79.2} &  \textbf{79.6} &  37.7 &  \textbf{54.0} &  \textbf{52.6} &  \textbf{66.2} & \textbf{68.8}

\\
\midrule
\multicolumn{10}{c}{{\small{Supervision With \hl{\textbf{``5K Unique''}} Seed Dictionary}}} \\
\midrule

\multicolumn{6}{l}{\textbf{Sup.\blue{/Semi-sup.} Baselines}}  \\

\newcite{artetxe2017acl} & 81.3 & 83.3 & 72.8 & 72.6 & 76.3 & 77.6 & 24.1 & 45.3 & 47.5 & 60.3 & 64.1\\

\newcite{artetxe2018aaai} & 80.8 & 84.5 & 73.3 & 74.3 & 77.4 & 79.7 & 42.0 & 54.7 & 51.5 & {68.2} & 68.7\\

\newcite{conneau2018word} & 81.6 & 83.5 & 74.1 & 72.7 & 77.8 & 77.2 & 34.3 & 48.5 & 49.0 & 60.7 &  66.0\\

\newcite{joulin-2018} & \textbf{83.4} & {85.4} & \textbf{77.0} & \textbf{76.4} & {78.7} & \textbf{81.6} & \textbf{41.3} & 54.0 & \textbf{58.1} & 67.4 & \textbf{70.4}\\
 
\newcite{jawanpuria2018learning} & \blue{81.3} & \textbf{{\blue{86.3}}} & \blue{74.5} & \blue{75.9} & \blue{78.6} & \blue{81.3} & \blue{38.7} & \blue{53.4} & \blue{52.3} & \blue{67.6} & 68.9\\
 
\newcite{patra-19}  & \blue{82.2} & \blue{84.6} & \blue{75.6} & \blue{73.7} & \blue{77.8} & \blue{78.6} & \blue{35.0} & \blue{51.9} & \blue{52.2} & \blue{65.2} & 69.5\\



\lnmap & 80.9 &  80.8 &  74.9 &  72.3 &  77.1 &  76.5 &  40.7 &  56.6 &  52.2 &  64.8 & 67.7
\\

\lnmaplinb & \textbf{83.4} &  85.7 &  75.5 &  75.4 &  \textbf{79.0} &  81.1 &  39.5 &  \textbf{56.8} &  53.8 &  \textbf{68.4} & 69.9

\\
\midrule
\multicolumn{10}{c}{{\small{Supervision With \hl{\textbf{``5K All''}}(5K Unique Source Words) Seed Dictionary}}} \\
\midrule

\multicolumn{6}{l}{\textbf{Sup.\blue{/Semi-sup.} Baselines}}  \\

\newcite{artetxe2017acl} & 81.2 & 83.5 & 72.8 & 72.5 & 76.0 & 77.5 & 24.4 & 45.3 & 47.3 & 61.2 & 64.2\\

\newcite{artetxe2018aaai} & 80.5 & 83.8 & 73.5 & 73.5 & 77.1 & 79.2 & 41.2 & 55.5 & 50.5 & 67.3  & 68.2\\

\newcite{conneau2018word} & 81.6 & 83.2 & 73.7 & 72.6 & 77.3 & 77.0 & 34.1 & 49.4 & 49.8 & 60.7 & 66.0\\

\newcite{joulin-2018} & \textbf{84.4} & \textbf{86.4} & \textbf{79.0} & {76.0} & {79.0} & {81.4} & \textbf{42.2} & 55.5 & \textbf{57.4} & 67.0  & \textbf{70.9}\\
 
\newcite{jawanpuria2018learning} & \blue{81.4} & \blue{85.5} & \blue{74.7} & \textbf{\blue{76.7}} & \blue{77.8} & \blue{80.9} & \blue{38.1} & \blue{53.3} & \blue{51.1} & \blue{67.6}  & 68.7\\
 
\newcite{patra-19}  & \blue{84.0} & \textbf{\blue{86.4}} & \blue{78.7} & \blue{76.4} & \textbf{\blue{79.3}} & \textbf{\blue{82.4}} & \blue{41.1} & \blue{53.9} & \blue{57.2} & \blue{64.8} & 70.4\\



\lnmap & 80.5 &  82.2 &  73.9 &  72.7 &  76.7 &  78.3 &  41.5 &  57.1 &  53.5 &  67.1 & 68.4
\\

\lnmaplinb & 82.9 &  \textbf{86.4} &  75.5 &  75.9 &  78.1 &  81.4 &  39.3 &  \textbf{57.3} &  52.3 &  \textbf{67.8} & 69.6

\\
\bottomrule
\end{tabular}}
\caption{Word translation accuracy (P@1) on \textbf{resource-rich} languages on \textbf{MUSE dataset} using {fastText}.} 
\label{tab:high-resource-results} 
\end{table*}
\endgroup

\subsection{Results on Resource-rich Languages} \label{subsec:high-resource-con-dinu}


Table \ref{tab:high-resource-results} shows the results for 5 resource-rich language pairs (10 translation tasks) from the MUSE dataset. We notice that our model achieves the highest accuracy in all the tasks for  {``1K Unique''}, 4 tasks for  {``5K Unique''}, 3 for {``5K All''}.



We show the results on the VecMap dataset in Table \ref{tab:dinu-results}, where there are 3 resource-rich language pairs, and one low-resource pair (En-Fi) with a total of 8 translation tasks. Overall, we have similar observations as in MUSE -- our model outperforms other models on 7 tasks for {``1K Unique''}, 4 tasks for {``5K Unique''}, and 4 for {``5K All''}.




\begingroup
\begin{table}[t!]
\centering
\small
\setlength{\tabcolsep}{4pt}
\scalebox{0.7}{\begin{tabular}{l|cc|cc|cc|cc||c}
&\multicolumn{2}{c}{\textbf{En-Es}}  &        \multicolumn{2}{c}{\textbf{En-It}} &\multicolumn{2}{c}{\textbf{En-De}}& \multicolumn{2}{c}{\textbf{En-Fi}} &
\textbf{\blue{Avg.}}
\\
&\textbf{$\rightarrow$} & \textbf{$\leftarrow$} & \textbf{$\rightarrow$} & \textbf{$\leftarrow$} & 
\textbf{$\rightarrow$} & \textbf{$\leftarrow$} & 
\textbf{$\rightarrow$} & \textbf{$\leftarrow$} 
\\       
\midrule

\multicolumn{9}{l}{\textbf{Unsupervised Baselines}} \\

\newcite{Artetxe-2018-acl} & 36.9 & 31.6 & 47.9 & 42.3 & 48.3 & 44.1 & 32.9 & 33.5 & 39.7\\

\newcite{conneau2018word} & 34.7 & 0.0 & 44.9 & 38.7 & 0.0 & 0.0 & 0.0 & 0.0 & 14.8\\

\newcite{mohi-joty:2019} & \blue{37.4} & \blue{31.9} & \blue{47.6} & \blue{42.5} & \blue{0.0} & \blue{0.0} & \blue{0.0} & \blue{0.0}  & 19.9\\

\midrule
\multicolumn{10}{c}{{\small{Supervision With \hl{\textbf{``1K Unique''}} Seed Dictionary}}} \\
\midrule

\multicolumn{9}{l}{\textbf{Sup.\blue{/Semi-sup.} Baselines}}  \\

\newcite{artetxe2017acl}  & 33.3 & 27.7 & 43.9 & 38.1 & 46.8 & 40.8 & 30.4 & 26.0 & 35.9\\

\newcite{artetxe2018aaai} &  29.0 & 20.0 & 38.6 & 29.2 & 36.3 & 26.0 & 25.8 & 15.0 & 27.5\\

\newcite{conneau2018word} & {35.7} & {30.8} & {45.4} & 38.3 & {46.9} & {42.3} & {29.1} & 27.2 & 37.0\\

\newcite{joulin-2018} & 24.2 & 17.9 & 33.9 & 25.1 & 31.6 & 25.5 & 21.9 & 14.5 & 24.4\\
 
\newcite{jawanpuria2018learning}  & \blue{31.5} & \blue{23.2} & \blue{39.2} & \blue{32.4} & \blue{39.1} & \blue{30.9} & \blue{26.8} & \blue{21.4} & 30.6\\ 

\newcite{patra-19}   & \blue{31.4} & \blue{30.5} & \blue{30.9} & \blue{38.8} & \textbf{\blue{47.9}} & {\blue{43.7}} & {\blue{30.5}} & {\blue{31.6}} & 35.7\\



\lnmap & 32.9 &  28.6 &  44.2 &  39.1 &  43.0 &  39.2 &  26.6 &  25.4 & 34.9
\\

\lnmaplinb & \textbf{36.5} & \textbf{ 33.6} &  \textbf{46.0} &  \textbf{40.1} &  46.4 &  \textbf{44.8} &  \textbf{31.7} &  \textbf{37.1} & \textbf{39.5}

\\
\midrule
\multicolumn{10}{c}{{\small{Supervision With \hl{\textbf{``5K Unique''}} Seed Dictionary}}} \\
\midrule

\multicolumn{9}{l}{\textbf{Sup.\blue{/Semi-sup.} Baselines}}  \\

\newcite{artetxe2017acl} & 33.3 & 27.6 & 43.9 & 38.4 & 46.0 & 41.1 & 30.9 & 25.7 & 35.9\\

\newcite{artetxe2018aaai} & \textbf{37.6} & {34.0} & 45.7 & \textbf{41.6} & {47.2} & {45.0} & {34.0} & \textbf{38.8} & 40.2\\

\newcite{conneau2018word} & 36.0 & 31.1 & 46.0 & 38.8 & 47.6 & 43.2 & 31.1 & 28.2 & 37.8\\

\newcite{joulin-2018} & 34.2 & 31.1 & 43.1 & 37.2 & 44.5 & 41.9 & 30.9 & 34.7 & 37.2\\
 
\newcite{jawanpuria2018learning}   & \blue{36.9} & \blue{33.3} & \textbf{{\blue{47.1}}} & \blue{39.9} & {\blue{47.7}} & \blue{44.6} & \textbf{\blue{35.1}} & \blue{38.0} & 40.2\\ 

\newcite{patra-19}   & \blue{34.3} & \blue{31.6} & \blue{41.1} & \blue{39.3} & \blue{47.5} & \blue{43.6} & \blue{30.7} & \blue{33.4} & 37.7\\



\lnmap & 33.4 &  27.3 &  44.1 &  38.9 &  42.5 &  39.4 &  29.7 &  28.6 & 35.5
\\

\lnmaplinb & 37.1 &  \textbf{34.1} &  46.2 &  40.3 &  \textbf{47.7} &  \textbf{45.6} &  33.3 & \textbf{ 38.8} & \textbf{40.3}

\\
\midrule
\multicolumn{10}{c}{{\small{Supervision With \hl{\textbf{``5K All''}} (5K Unique Source Words) Seed Dictionary}}} \\
\midrule

\multicolumn{9}{l}{\textbf{Sup.\blue{/Semi-sup.} Baselines}}  \\

\newcite{artetxe2017acl} & 32.7 & 28.1 & 43.8 & 38.0 & 47.4 & 40.8 & 30.8 & 26.2 & 36.0\\

\newcite{artetxe2018aaai} & {38.2} & {33.4} & {47.3} & \textbf{41.6} & {47.2} & {44.8} & \textbf{34.9} & {38.6} & \textbf{40.8}\\

\newcite{conneau2018word} & 36.1 & 31.2 & 45.7 & 38.5 & 47.2 & 42.8 & 31.2 & 28.3 & 37.7\\

\newcite{joulin-2018} & 35.5 & 31.2 & 44.6 & 37.6 & 46.6 & 41.7 & 32.1 & 34.4 & 38.0\\

\newcite{jawanpuria2018learning} & \blue{37.5} & \blue{33.1} & \textbf{\blue{47.6}} & \blue{40.1} & \textbf{\blue{48.8}} & {\blue{45.1}} & \blue{34.6} & \blue{37.7} & 40.6\\ 

\newcite{patra-19}   & \blue{34.5} & \blue{32.1} & \blue{46.2} & \blue{39.5} & \blue{48.1} & \blue{44.1} & \blue{31.0} & \blue{33.6} & 39.4\\
 


\lnmap & 33.7 &  27.9 &  43.7 &  38.9 &  43.6 &  39.2 &  29.9 &  31.5 & 36.1
\\

\lnmaplinb & \textbf{37.8} &  \textbf{34.6} &  46.7 &  40.2 &  47.7 &  \textbf{45.2} &  34.1 &  \textbf{38.9} & 40.6

\\
\bottomrule
\end{tabular}}
\caption{Word translation accuracy (P@1) on \textbf{VecMap dataset} using {CBOW} embeddings.} 
\label{tab:dinu-results} 
\end{table}
\endgroup

\subsection{Effect of Non-linearity in Autoencoders}
\label{subsec:lin-ae}

The comparative results between our model variants in Tables \ref{tab:low-resource-results} - \ref{tab:dinu-results} reveal that \lnmap\ (with nonlinear autoencoders) works better for low-resource languages, whereas \lnmaplin\ works better for resource-rich languages. This can be explained by the geometric similarity between the embedding spaces of the two languages. 



In particular, we measure the geometric similarity of the language pairs using the \textbf{Gromov-Hausdorff (GH)} distance \cite{patra-19}, which is recently proposed to quantitatively estimate isometry between two embedding spaces.\footnote{\href{https://github.com/joelmoniz/BLISS}{https://github.com/joelmoniz/BLISS}} 
From the measurements (Tables \ref{tab:low-resource-results}-\ref{tab:high-resource-results}), we see that etymologically {close language pairs} have lower {GH distance} compared to etymologically distant and low-resource language pairs.\footnote{We could not compute GH distances for the VecMap dataset; the metric gives `inf' in the  BLISS framework.} Low-resource language pairs' high {GH distance} measure implies that English and those languages embedding spaces are far from isomorphism. {Hence, we need strong non-linearity for those distant languages.}

\subsection{Dissecting \lnmap\ }
\label{subsec:dissection}

We further analyze our model by dissecting it and measuring the contribution of its different components. Specifically, our goal is to assess the contribution of back-translation, reconstruction, non-linearity in the mapper, and non-linearity in the autoencoder. We present the ablation results in Table \ref{tab:ablation-results} on 8 translation tasks from 4 language pairs consisting of 2 resource-rich and 2 low-resource languages. We use MUSE dataset for this purpose. All the experiments for the ablation study are done using {``1K Unique''} seed dictionary. 

\vspace{-0.5em}
\paragraph{$\ominus$ Reconstruction loss:} For removing the reconstruction loss from the full model, on average high-resource language pairs lose accuracy by 0.9\% and 5.3\% for from and to English, respectively. The losses are even higher for low-resource language pairs,  on average 2.5\% and 6.4\% in accuracy.


\begingroup
\setlength{\tabcolsep}{3pt}
\begin{table}[t!]
\centering
\small
\scalebox{0.8}{\begin{tabular}{l|cc|cc||cc|cc}
& \multicolumn{4}{c||}{\textbf{Resource-rich}}  &        \multicolumn{4}{c}{\textbf{Low-Resource}} \\
\midrule

&\multicolumn{2}{c}{\textbf{En-Es}}  &        \multicolumn{2}{c||}{\textbf{En-It}}&
\multicolumn{2}{c|}{\textbf{En-Ta}}& 
\multicolumn{2}{c}{\textbf{En-Bn}} 
\\
&\textbf{$\rightarrow$} & \textbf{$\leftarrow$} & \textbf{$\rightarrow$} & \textbf{$\leftarrow$} & 
\textbf{$\rightarrow$} & \textbf{$\leftarrow$} & 
\textbf{$\rightarrow$} & \textbf{$\leftarrow$} 
\\       
\midrule
\lnmapb & 80.1 & 80.2 & 77.1 & 75.3 & 17.6 & 21.2 & 18.4 & 27.2\\
\midrule
$\ominus$ Recon. loss & 79.6 & 75.4 & 75.7 & 69.4 & 14.8 & 14.9 & 16.2 & 20.7 \\
$\ominus$ Back-tran. loss & 79.8 & 79.1 & 76.6 & 74.4 & 16.7 & 20.3 & 16.5 & 26.7\\
\midrule
$\oplus$ Linear mapper & 78.8 & 78.9 & 76.3 & 74.7 & 16.6 & 20.2 & 18.0 & 26.3\\
$\oplus$ Procrustes sol. & 75.9 & 73.9 & 72.0 & 72.2 & 11.1 & 12.1 & 12.2 & 14.8\\
\midrule
$\oplus$ Linear autoenc. & 83.2 & 85.5 & 79.2 & 79.6 & 14.5 & 20.3 & 16.5 & 26.1\\

\bottomrule
\end{tabular}}
\caption{Ablation study of \lnmap\ with ``1K Unique" dictionary. $\ominus$ indicates the component is removed from the full model, and `$\oplus$' indicates the component is added by replacing the corresponding component.}
\label{tab:ablation-results} 
\end{table}

\endgroup

\vspace{-0.5em}
\paragraph{$\ominus$ Back-translation (BT) loss:} Removing the BT loss also has a negative impact, but not as high as the reconstruction. This is because the reconstruction loss  (Eq. \ref{reconsAloss}) also covers the BT signal.


\vspace{-0.5em}
\paragraph{$\oplus$ Linear mapper:} If we replace the non-linear mapper with a linear one in the full model, we see that the effect is not that severe. The reason can be explained by the fact that the autoencoders are still non-linear, and the non-linear signal passes through back-translation and reconstruction.

 \vspace{-0.5em}
\paragraph{$\oplus$ Procrustes solution:} To assess the proper effect of the non-linear mapper, we need to replace it with a linear mapper through which no non-linear signal passes by during training. This can be achieved by replacing the non-linear mapper with the Procrustes solution. The results show an adverse effect on removing non-linearity in the mapper in all the language pairs. However, low-resource pairs' performance drops quite significantly.

\vspace{-0.5em}
\paragraph{$\oplus$ Linear autoencoder:} For high-resource language pairs, linear autoencoder works better than the non-linear one. However, it is the opposite for the low-resource pairs, where the performance drops significantly for the linear autoencoder. 



\section{Conclusions}
\label{sec:conclusion}

We have presented a novel semi-supervised framework \lnmap\  to learn the cross-lingual mapping between two monolingual word embeddings. Apart from exploiting weak supervision from a small (1K) seed dictionary, our \lnmap\ leverages the information from monolingual word embeddings. {In contrast to the existing methods that directly map word embeddings using the isomorphic assumption, our framework is independent of any such strong prior assumptions. \lnmap\ first learns to transform the embeddings into a latent space and then uses a non-linear transformation to learn the mapping.} To guide the non-linear mapping further, we include constraints for back-translation and original embedding reconstruction.

Extensive experiments with {fifteen different language pairs} comprising high- and low-resource languages show the efficacy of non-linear transformations, especially for low-resource and distant languages. Comparison with existing supervised, semi-supervised, and unsupervised baselines show that \lnmap\ learns a better mapping. With an in-depth ablation study, we show that different components of \lnmap\ works in a collaborative nature. 

\section*{Acknowledgments}

We would like to thank the anonymous reviewers for their helpful comments. Shafiq Joty would like to thank the funding support from NRF (NRF2016IDM-TRANS001-062), Singapore.

\bibliographystyle{acl_natbib}
\bibliography{word-tr.bib}

\appendix
\section{Appendix}

\subsection{Reproducibility Settings}
\begin{itemize}
    \item Computing infrastructure - Linux machine with a single GTX 1080 Ti GPU
    \item PyTorch version 1.2.0
    \item CUDA version 10.0
    \item cuDNN version 7.6.0
    \item Average runtime - 15-20 minutes
\end{itemize}

\subsection{Optimal Hyperparameters}

\begin{table}[H]
\begin{center}
\begin{tabular}{l|c}
\hline 
\bf Hyperparameter & \bf Value \\  
\hline
\multicolumn{2}{l}{\underline{Encoder}}\\
\#layers & 3\\
input dim & 300\\
hidden dim & 350-400\\
output dim & 350-400\\
hidden non-linearity & \texttt{PReLU}\\
output non-linearity & \texttt{linear}\\

\hline
\multicolumn{2}{l}{\underline{Decoder}}\\
\#layers & 3\\
input dim & 350-400\\
hidden dim & 350-400\\
output dim & 300\\
hidden non-linearity & \texttt{PReLU}\\
output non-linearity & \texttt{tanh}\\
\hline
\end{tabular}
\end{center}
\caption{\label{autoenc-hyperparameters} Optimal hyper-parameter settings for autoencoder.}
\end{table}

\begin{table}[H]
\begin{center}
\begin{tabular}{l|c}
\hline 
\bf Hyperparameter & \bf Value \\  
\hline
type & \texttt{linear/non-linear}\\
\#layers & 2\\
input dim & 350-400\\
hidden dim & 400\\
output dim & 350-400\\
hidden non-linearity & \texttt{tanh}\\
output non-linearity & \texttt{linear}\\
\hline
\end{tabular}
\end{center}
\caption{\label{autoenc-hyperparameters} Optimal hyper-parameter settings for mapper.}
\end{table}

\begin{table}[H]
\begin{center}
\begin{tabular}{l|c}
\hline 
\bf Hyperparameter & \bf Value \\  
\hline
normalization & \texttt{renorm,center,renorm}\\
\#iterations & \texttt{dynamic}\\
sup. dict size & 1K-5K\\
batch size & 128 \\
autoenc. epochs & 25\\
mapper epochs & 100\\
nearest-neighbor & \texttt{CSLS}\\
autoenc. optimizer & \texttt{SGD}\\
autoenc. learning-rate & {0.0001}\\
mapper optimizer & \texttt{SGD}\\
mapper learning-rate & {0.0001}\\
mapping-loss weight & 1.0\\
cycle-loss weight & 1.0\\
recons.-loss weight & 1.0\\
\hline
\end{tabular}
\end{center}
\caption{\label{autoenc-hyperparameters} Optimal hyper-parameter settings for \lnmap\ training.}
\end{table}

\end{document}